\pdfoutput=1

\documentclass{article} 
\usepackage[preprint]{colm2025_conference}

\usepackage{microtype}
\usepackage[breaklinks]{hyperref}
\usepackage{url}
\usepackage{booktabs}

\usepackage{lineno}

\usepackage{graphicx}

\usepackage{wrapfig}
\usepackage{breakurl}

\usepackage{times}
\usepackage{latexsym}

\usepackage[T1]{fontenc}

\usepackage[utf8]{inputenc}

\usepackage{microtype}

\usepackage{inconsolata}

\usepackage{graphicx}

\usepackage{outlines}
\usepackage{todonotes}
\usepackage{natbib}
\usepackage{float}
\usepackage{booktabs}
\usepackage{amsmath}
\usepackage{cleveref}
\usepackage{array}       
\usepackage{multirow}    

\definecolor{darkblue}{rgb}{0, 0, 0.5}
\hypersetup{colorlinks=true, citecolor=darkblue, linkcolor=darkblue, urlcolor=darkblue}

\title{Towards Developmentally Plausible Rewards: Communicative Success as a Learning Signal for Interactive Language Models}

\author{Lennart St\"opler* \\
Department of Computer Science\\
Heidelberg University \\
Heidelberg, Germany \\
\texttt{lennart.stoepler@uni-heidelberg.de} \\
\And
Rufat Asadli*, Ryan Cotterell \\
Department of Computer Science \\
ETH Z\"urich \\
Z\"urich, Switzerland \\
\texttt{\{rasadli,rcotterell\}@inf.ethz.ch} \\
\AND
Mitja Nikolaus \\
CerCo, CNRS, Toulouse \\
Toulouse, France \\
\texttt{mitja.nikolaus@cnrs.fr} \\
\AND
Alex Warstadt \\
University of California San Diego \\
San Diego, USA \\
\texttt{awarstadt@ucsd.edu}
}

\begin{document}

\ifcolmsubmission
\linenumbers
\fi

\maketitle

\begin{abstract}
We propose a method for training language models in an interactive setting inspired by child language acquisition. 
In our setting, a speaker attempts to communicate some information to a listener in a single-turn dialogue and receives a reward if communicative success is achieved.
Unlike earlier related work using image--caption data for interactive reference games, we operationalize communicative success in a more abstract language-only question--answering setting.
First, we present a feasibility study demonstrating that our reward provides an indirect signal about grammaticality.
Second, we conduct experiments using reinforcement learning to fine-tune language models.
We observe that cognitively plausible constraints on the communication channel lead to interpretable changes in speaker behavior.
However, we do not yet see improvements on linguistic evaluations from our training regime. 
We outline potential modifications to the task design and training configuration that could better position future work to use our methodology to observe the benefits of interaction on language learning in computational cognitive models.
\end{abstract}

\section{Introduction}

Language is useful to humans because it empowers us to interact with other agents.
We can use language to make requests, entertain one another, and---chiefly---to exchange information \citep{searleSpeechActsEssay1969}.
Language is so useful in human life that new ones spontaneously arise in communities without a common language \citep{senghas2001children,kouwenberg2009handbook}.
Infants and children, therefore, have an incredible motivation to acquire language, with their successes and failures to do so routinely reflected back to them through the responses of others \citep{clarkConversationLanguageAcquisition2018,clarkConversationalRepairAcquisition2020}. 

By contrast, neural language models \citep[LMs;][]{elman1990finding,openai2022chatgpt,ai@meta2024llama} derive no utility from communication.
These models have recently become ubiquitous due to their remarkable ability to acquire human language \citep{linzen2019can,warstadt2022what,frank_bridging_2023}, but unlike humans, their only objective is to predict the next token in a text authored by some other agent.%
\footnote{We focus our discussion on base models, not models fine-tuned on human preferences \citep{RLHF}, as preference tuning is typically invoked to improve LMs' usability as products rather than cognitive models.}
While this clearly has not hindered LMs' utility to human users, it is a problem for (at least) two reasons:
First, LMs have the potential to transform computational modeling of human language processing and acquisition \citep{warstadt2022what}, but they are currently too divergent from humans to meet their full potential \citep{baroni2022proper}.
Second, humans are orders of magnitude more data-efficient at acquiring language than LMs \citep{babylm}, and this may be due in part to the presence of an interactive learning signal.

\begin{figure}[t]
\centering
\includegraphics[width=0.8\columnwidth]{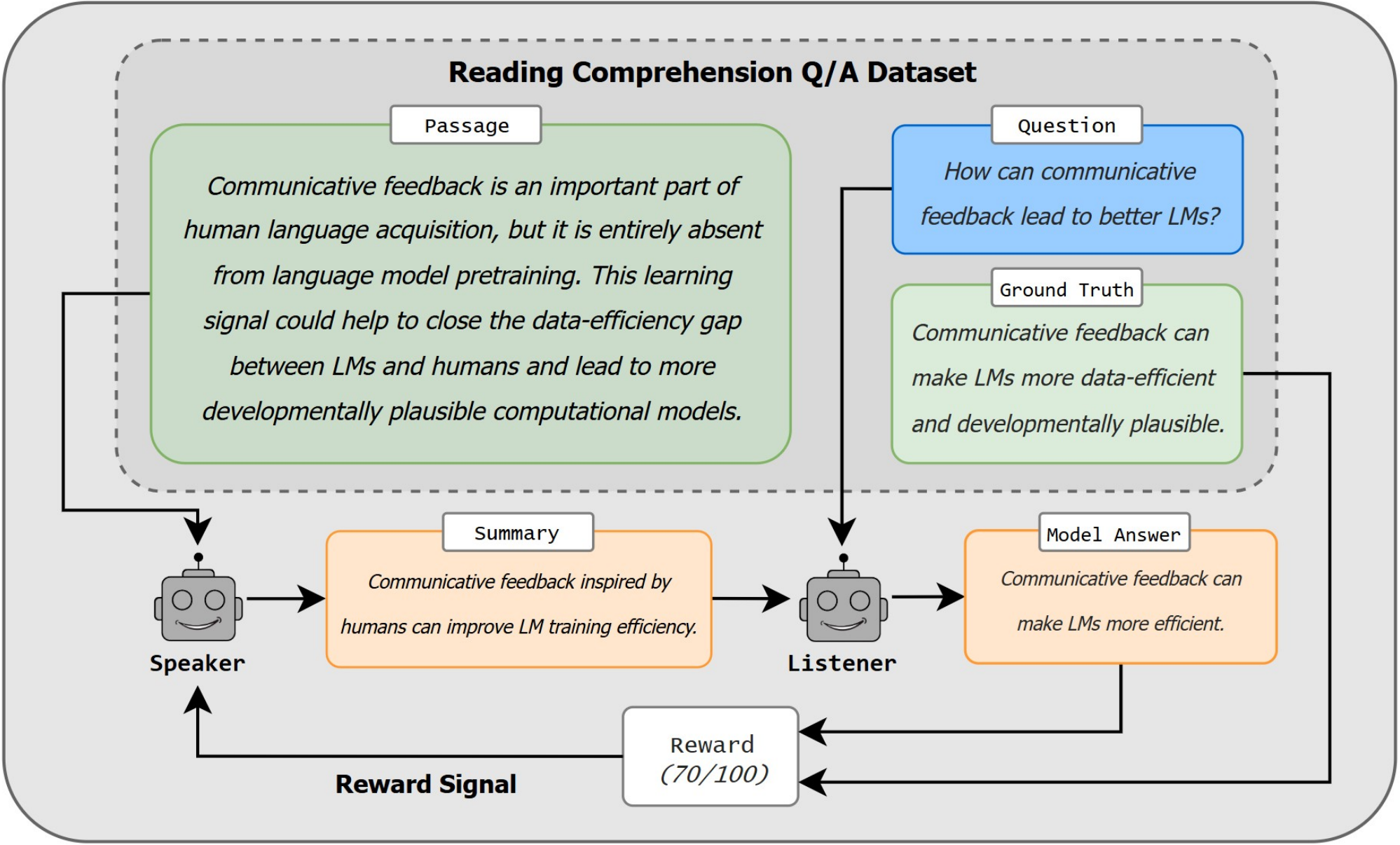}
\vspace{-1em}
\caption{Illustration of the summarization game.}\label{fig:setup}
\end{figure}

We introduce a novel training regime for LMs that incorporates interactive learning.
In our setting, illustrated in \cref{fig:setup}, a speaker LM simulating a child, learns from interacting with a mature listener LM and observing its degree of communicative success.
This setup is closely related to prior work using reinforcement learning to train LMs to perform reference games \citep{lazaridou2020multiagentmeets,nikolaus-fourtassi-2021-modeling,rita2024language} with two main exceptions:
(1) It is designed for acquisition of natural language as opposed to emergent communication protocols \citep{rita2024language}; 
and (2) it takes place in a language-only world as opposed to a vision--language world \citep{lazaridou2020multiagent,nikolaus-fourtassi-2021-modeling}, enabling the use of simpler models and more complex linguistic input.

The present contribution is a proof of concept.
After considering major design choices in the training scenario, we conduct a feasibility study in which we show that our notion of communicative success carries some learning signal for grammaticality.
In particular, we find that communicative success degrades when the listener receives ungrammatical input.
Next, we conduct experiments training T5 \citep{t5} from scratch and in a fine-tuning setting.
We systematically explore ways to operationalize cost in production or comprehension based on the length or surprisal of the speaker's output.
We find that a length-based bottleneck incentivizes ``telegraphic'' outputs lacking function words and causes LM acceptability judgments to degrade, while a surprisal-based bottleneck leads to outputs with grammatical properties similar to that of the original LM.
None of our experiments lead to improvements in models' grammatical ability.
However, we see considerable opportunity to improve this methodology and further explore its implications for natural language processing and cognitive modeling.

\section{Background}

\subsection{Developmentally Plausible Language Models}

Thanks to their remarkable ability to acquire human language, language models are increasingly being explored for their potential as cognitive models or models of child language acquisition \citep{elman1996rethinking,linzen2019can,warstadt2022what,wilcox2023using,milliereforthcominglanguage}.
While there is some controversy about LMs' validity as such \citep[see, e.g.][]{baroni2022proper,katzir2023why,lan2024large}, there is little disagreement that \emph{the way} in which LMs acquire language is vastly divergent from how children do so \citep{babylm}.
As there are too many differences to elaborate on in detail,\footnote{For instance, LMs are trained in a text-only environment without access to speech, vision, and other sensory modalities. The input to LMs often comes from written and edited domains. The Transformer architecture on which most LMs are based is not neurologically motivated. See \citet{warstadt2022what} for further discussion. 
} we will discuss the two that are most relevant to the present work: the training objective, and data efficiency.

\paragraph{The Training Objective}
LM training objectives fail to capture the learning signals that humans use to acquire language.
The objective function that autoregressive language models like GPTs, Llamas, and Pythias \citep{openai2022chatgpt,ai@meta2024llama,pmlr-v202-biderman23a} optimize diverges from human learning in (at least) two respects: 
(a) the exclusive reliance on the next-token prediction task and 
(b) the use of teacher forcing.

Prediction tasks in general may indeed be cognitively \emph{plausible}.
Predictive coding theory, originally developed to explain visual processing in humans \citep{srinivasan1982predictive,huang2011predictive} but increasingly applied to language processing \citep{lupyan2015words,shain2020fmri}, holds that the brain streamlines processing by predicting upcoming sensory input.
Thus, the next-token prediction task used to train autoregressive LMs \citep{elman1990finding} is not necessarily problematic from the standpoint of cognitive modeling.

However, prediction is clearly not the only learning signal available to human learners.
Most relevant to this work, and as we review in \cref{sec:human_interaction}, children also learn through interaction with their caregivers and other children, by optimizing for communicative success, or through corrective feedback \citep{brunerChildTalkLearning1985,ninioLanguageAcquisitionLanguage1988,tomaselloConstructingLanguageUsageBased2003,yurovskyCommunicativeApproachEarly2018,clarkConversationLanguageAcquisition2018,nikolaus_communicative_2023}.

Furthermore, language models are trained through teacher-forcing, in which predictions are limited to a single unit in the future, and feedback is given at each unit.
The unit of prediction in LM training is typically subword tokens, which have no grounding in linguistic structure.
The human brain could conceivably learn from prediction across the linguistic hierarchy, from phones to entire utterances.
Ultimately, the objective function for human language combines (among other things) interactive and predictive learning signals over varying time scales.

\paragraph{Data Efficiency}

Another limitation of LMs as models of human learning is their need for massive training corpora.
Today's large LMs like Llama \citep{ai@meta2024llama} are trained on about 10 trillion words of input.
By comparison, most children are exposed to no more than 100 million words by age 12 \citep{gilkerson2017mapping}, and LMs trained on developmentally plausible data quantities do not show the same linguistic proficiency as children \citep{zhang2021when}.
Closing this data-efficiency gap is now an area of active research \citep{babylm}, with approaches ranging from curriculum learning \citep{babylm2023climb} to novel neural architectures \citep{babylm2023notall}.
However, interaction has hardly been explored as a technique for this purpose (see \cref{sec:lm_interaction} for a review).
It stands to reason that the additional learning signals that children receive from interaction could account for some portion of the data-efficiency gap.

\subsection{Interaction in Human Language Learning}\label{sec:human_interaction}

Many debates among researchers in language acquisition have revolved around the importance of interaction and feedback signals for language learning \citep{goldLanguageIdentificationLimit1967,saxtonContrastTheoryNegative1997,brownDerivationalComplexityOrder1970,marcusNegativeEvidenceLanguage1993,chouinardAdultReformulationsChild2003,schonebergerThreeMythsLanguage2010,clarkConversationalRepairAcquisition2020}.
While these studies have led to mixed results, more recent analyses performed on a large scale highlight the potential of communicative signals \citep{hillerDatadrivenInvestigationCorrective2016,nikolaus_communicative_2022,nikolaus_communicative_2023a}.

According to the typology proposed in \citet{nikolaus_communicative_2023}, interactive learning signals can take varying degrees of explicitness, from explicit acknowledgments, clarification requests, corrections, and reformulations to more implicit signals such as response contingency. 
In practice, communicative feedback signals can take the form of words and sentences (e.g. ``What did you say?'') or be communicated non-verbally (e.g. a head shake to communicate lack of understanding).
This typology also encompasses a direct link between the valence of feedback and communicative success/failure, which we will exploit in this work:
If a communicative attempt was successful, the speaker receives positive feedback. In case of communication failure or breakdown, negative feedback is received, encouraging the speaker to adapt and improve their communication strategy for subsequent utterances.\looseness=-1


\subsection{Interactive LM training}\label{sec:lm_interaction}

Explicit computational models of language acquisition incorporating communicative feedback are a relatively new phenomenon. \citet{nikolaus-fourtassi-2021-modeling} first studied its effect on language acquisition by combining cross-situational learning and reinforcement learning. However, there is a breadth of research from adjacent fields that informs this work.\looseness=-1

In \citet{population}, a population of speaker and listener models is grounded in the source language through supervised learning and then taught to cooperate on referential tasks, including text-only ones. However, in their setup both the speaker and listener start from scratch and are jointly trained, whereas our model of language acquisition assumes an asymmetric communication between a weaker speaker model and a fluent listener.\looseness=-1

Other work from language emergence uses functional reference games (e.g. Lewis signaling games \citep{Lewis1969-LEWCAP-4}), with some experiments also involving fixed listener agents \citep[e.g.][]{lazaridou2020multiagentmeets, lowe2020interaction, liu2023computational,jacob2021multitasking}. 
These setups rely on multi-modal data in the form of images or simulated environments to serve as a semantic grounding for communication in order to evaluate communicative success.
While in principle the inclusion of multimodal input is a desirable factor when modeling human language acquisition, in practice the multimodal datasets used in past studies are of limited size and linguistic complexity.


If speaker and listener models are jointly trained, the semantics of their communication likely diverge from the source language - a process known as semantic drift \citep{jacob2021multitasking, lazaridou2020multiagentmeets, lu2020countering}. 
Using a fixed listener largely prevents semantic drift as it anchors communication to the original training data. However, \citet{lazaridou2020multiagentmeets} show that drift can still occur if the speaker learns to exploit statistical anomalies of the listener's behavior.


Recent work related to preference alignment - like Reinforcement Learning from Human Feedback \citep{RLHF} - uses very similar training methods to our work. However, the motivation is vastly different. These methods focus on fine-tuning existing models to better align with user preferences, whereas we attempt to validate a cognitive model of communicative feedback. Nevertheless, this line of work has spawned a collection of text-based communication tasks \citep{abdulhai2023lmrl}, which, though originally intended for model fine-tuning, could provide a valuable resource for future work on communicative feedback.

Finally, since the listener is a fully trained model, this work can be cast as an instance of reinforcement-based knowledge distillation \citep{xu2024survey}. However, unlike model distillation, our task is not designed to be effective at approximating the behavior of the listener model. Instead, we focus on the cognitive plausibility of the setup.




\section{The Abstract Reference Game}\label{methods}

As in prior work (see above) we model learning from communicative feedback by applying reinforcement learning to a \textbf{reference game} (i.e. a Lewis game). 
In a reference game, a speaker must produce a message that will allow a listener to identify a specific referent---often, selecting a target image from an array of images \citep{Lewis1969-LEWCAP-4}.
Our approach, which we refer to as the \textbf{abstract reference game} is novel in that we recast this setup in a language-only world.
While, ultimately, interaction in a multimodal world will lead to the most developmentally plausible computational models \citep{vong2024grounded}, the linguistic information available in most current vision--language datasets is limited to mainly short concrete description of the visual content \citep[e.g.,][]{changpinyo2021conceptual}.
In order to expose model learners to abstract concepts, complex syntax, and rich rhetorical structure, we deem it necessary at present to include unimodal text data.\looseness=-1

The question remains: How does one play a reference game in a language-only world?
The agent's primary goal in a reference game, as in much of human discourse, is the exchange of information, i.e., the assertion propositions \citep{griceLogicConversation1975,stalnakerAssertion1978}.
Though propositions are abstract, they can be seen as referring to sets of possible worlds, namely those worlds in which the proposition is true \citep{carnap1947meaning,lewis1986plurality}.
Thus, in the abstract reference game, the goal of the speaker is to communicate a message that will allow the listener to identify a target set of possible worlds.




\subsection{The Summarization Game}



The way in which we operationalize the abstract reference game uses a combination of summarization \citep{el2021automatic} and question--answering (QA) \citep{khashabi2020unifiedqa} tasks common in natural language processing.%
\footnote{A similar idea was independently  proposed by \citet{giulianelli2022pragmatic} and implemented in unpublished work by \cite{BA}.}
A complete interaction proceeds in three parts:
First, in an information-seeking dialogue, there is an information asymmetry, where the speaker has some private knowledge that they want to communicate.
In our setting, the speaker is provided with prior knowledge in the form of a pre-written passage and must summarize the content to the listener while balancing competing goals of thoroughness and brevity.
Second, listener goals in information exchange can be formalized as a Question Under Discussion \citep{roberts1996informationa}, i.e. the (possibly implicit) question that the listener is trying to resolve.
In our setting, the listener is given a pre-written question about the passage and must generate an answer based solely on the speaker's summary.
Third, speakers and listeners routinely query and signal understanding through backchannels, nods, and other forms of communicative feedback \citep{schegloff1992repair}.
In our setting, the listener's comprehension is evaluated by comparing their answer to a ground-truth answer, and the speaker receives a reward reflecting its accuracy and completeness.
We re-purposed existing QA datasets (\cref{sec:implementation}) as the source of the passages, questions, and ground-truth answers.
Fig.~\ref{fig:setup} provides an illustration of our setup.


\subsection{Communication Bottleneck}\label{sec:bottleneck}

Communication between humans is constrained by practical limitations such as the effort a speaker expends in producing an utterance and the difficulty of a listener in processing or attending to the utterance. 
Furthermore, a naive implementation of the summarization is susceptible to a trivial strategy: 
A speaker could learn to reproduce the entire passage verbatim, allowing the listener to receive maximal information, thus transmitting maximal information with no need for syntactic or semantic knowledge.
We address this challenge and simulate human communicative constrains by introducing bottlenecks on the communication channel. 
While there are many conceivable ways to do so, we experiment with two bottlenecks: one based on length and the other based on surprisal under the listener model.\looseness=-1



\paragraph{Number of Tokens} 
The naive way to quantify the amount of information passed between speaker and listener is as the number of tokens the summary contains.
Indeed, shorter passages tend to have less information.
However, this need not be the case.
One can increase the information density of a text by removing words and phrases with little information.
We predict that this bottleneck has the potential to lead to the omission of function words, resulting in ungrammatical utterances or telegraphic speech.\looseness=-1

\paragraph{Surprisal} 
A more theoretically grounded approach uses surprisal \citep{shannon1948mathematical} to quantify information.
The surprisal $S$ of a sequence of tokens $\mathbf{x}$ is defined as the negative log probability of the sequence, which can be estimated using a language model: $S(\mathbf{x}) = -\log \mathrm{P}(\mathbf{x}) \approx -\log \mathrm{P}_{\mathrm{LM}}(\mathbf{x}).$


Surprisal has long been used as a proxy for comprehension difficulty \citep{levy2008expectationbased}.
Furthermore, we predict that a surprisal-based bottleneck can prevent the telegraphic speech heuristic, as surprisal is not only affected by the length of the summary but also the information density \citep{levy2007speakers} and grammaticality \citep{lau2017grammaticality}. 
Put differently, short texts packed with complex content or lacking function words would have high surprisal.\looseness=-1

A model's surprisal can be viewed as a proxy for the communication cost of that agent. 
Since both speaker and listener effort are relevant for optimized communication \citep{giulianelli2022pragmatic}  either (or both) would be ecologically valid. 
However, since we train the speaker, using its surprisal as a bottleneck has the potential problem of moving the goal post in ways that might encourage unnatural heuristics.
We therefore decided to consider only the listener surprisal as the bottleneck for the present work, and save an exploration of a speaker-based bottleneck for future work.

\paragraph{Cut-Off vs. Penalty} 
We conceive of two ways to apply a bottleneck, whether length- or surprisal based.
First, it could serve as a cutoff, truncating the summary once the limit is reached.
Second, it could be a penalty subtracted from the reward.
We adopt the second implementation as it yields a more continuous signal which we reason could lead to more stable training. 
We also train an unrestricted model without bottleneck as a baseline.

\subsection{Speaker and Listener Agents}
We choose to simulate a scenario in which the speaker is a language learner and the listener is a mature language user.
Practically, we achieve this by updating the weights for the speaker only, freezing the listener model.%
\footnote{An argument can be made that in a conversation all interlocutors adapt to each other. However, we judge this adaption to be negligible compared to the changes undergone during language acquisition.}
This has the added benefit of mitigating semantic drift \citep{lazaridou2020multiagentmeets}, as the listener agent cannot adapt to innovations in the protocol introduced by the speaker.
Both the speaker and the listener models are implemented as generative language models---in our case we use the encoder--decoder T5 model \citep{t5}. 
Generative models are necessary in order for the speaker to generate summaries and the listener to generate answers.



\section{Implementation}\label{sec:implementation}

\paragraph{Models and Data}
Both the speaker and listener agents are based on the T5-small architecture.%
\footnote{In a preliminary experiment, we found a measurable but small difference in performance and PPO convergence between T5-base (220M parameters) and T5-small (60M parameters).
We choose the smaller model in order to perform more runs and reduce variance due to random seed.}
For the speaker we use two versions of T5. One version is trained\footnote{Training is based on NanoT5 \citep{Nawrot2023nanoT5AP}. We follow them in training for 16.384 batches of 256 samples.} from scratch on a 70 million\footnote{Together with the 30 million tokens for the summarization game this leads to a cognitively plausible data budget of 100 million tokens.} token subset of C4 \citep{t5} to be used as bootstrapping. For the fine-tuning experiments, we use the weights of the pretrained model published in the original T5 paper \citep{t5}.
For the listener, we employ UnifiedQA \citep{khashabi2020unifiedqa}, a version of T5 fine-tuned on a combination of 17 QA datasets. 
During training from scratch we alternate between the same 70 million subset of C4 used for bootstrapping and the combined QA dataset published by \cite{khashabi2022unifiedqav2}. For the QA dataset we filter out all samples used in the training of UnifiedQA and subsample to 30 million tokens. 
As training data for fine-tuning, we use the validation set of SQuAD 2.0 \citep{squad2}.
The size of only 540,243 tokens enables fast convergence during training which we find necessary due to compute limitations.

\paragraph{Training Algorithm}
We optimize the speaker using Proximal Policy Optimization (PPO) \citep{PPO}. 
We perform an initial hyperparameter search fine-tuning the pretrained T5 model (see \cref{tab:hp_sweep} in the Appendix for search ranges), and select the best configuration (\cref{tab:hyperparameters}) which we use in all reported experiments. 

\paragraph{Reward and Penalty}
The reward\footnote{Terminology: The term ''reward'' in this setting has a double meaning. On the one hand, it refers to the ROUGE score between the predicted and the correct answer. On the other hand, it can mean the reward in the sense of reinforcement learning, i.e. the value used for a PPO optimization step. The two meanings coincide if no penalty is used, but are different otherwise. To avoid confusion we will use \textbf{reward} to describe the ROUGE score and refer to the other quantity as the \textbf{score}.} is implemented as the ROUGE-L F1-score \citep{lin-2004-rouge} between the answer produced by the listener and the ground truth. Unlike BLEU \citep{BLEU}, ROUGE has no minimum number of tokens. This is important, since some of the answers are very short, e.g. ``yes'' or ``no''. 

We implement surprisal- and length-based penalties.
They can be computed either as absolute values or relative to the full context's length and surprisal. We use relative values for all experiments. In either case, a hyperparameter $\lambda$ ($\lambda \in [0,1]$) is used to balance reward and penalty. For context $c$, summary $s$, predicted answer $a_p$, and ground truth answer $a_{gt}$ the scores used to update PPO are


\vspace{-0.8em}
\begin{equation*}
    \begin{aligned}[t]
        \text{SCORE}&_{\text{length}}(c, s, a_p, a_{gt}) :=
        & (1 - \lambda) * \text{ROUGE}(a_p, a_{gt}) - \lambda * \frac{|s|}{|c|},~\text{and}
    \end{aligned}
\end{equation*}
\vspace{-0.5em}
\begin{equation*}
    \begin{aligned}[t]
        \text{SCO}&\text{RE}_{\text{surprisal}}(c, s, a_p, a_{gt}) :=
        &(1 - \lambda) * \text{ROUGE}(a_p, a_{gt})
        - \lambda * \frac{\mathrm{S}_{\mathrm{L}}(s)}{\mathrm{S}_{\mathrm{L}}(c)},
    \end{aligned}
\end{equation*}

\vspace{-0.5em}
where $\mathrm{S}_{\mathrm{L}}(\cdot)$ is a short-hand notation for listener surprisal.

\paragraph{Instruction Prefix}
T5 is trained to receive an instruction prefix to differentiate downstream tasks. 
In preliminary experiments we compared the PPO training using a ``\texttt{Summarize:}'' prefix to training with no prefix at all.
Removing the prefix reduces the reward initially, but it quickly climbs to the same level obtained when training with a prefix. 
Since using no prefix reduces the model bias and allows greater exploration during reinforcement learning, we choose to not use a prefix for the main experiments.

\section{Experiment 1: Feasibility Study}\label{sec:feasibility}

\begin{wrapfigure}{r}{0.5\linewidth}
        \centering
        \includegraphics[width=\linewidth]{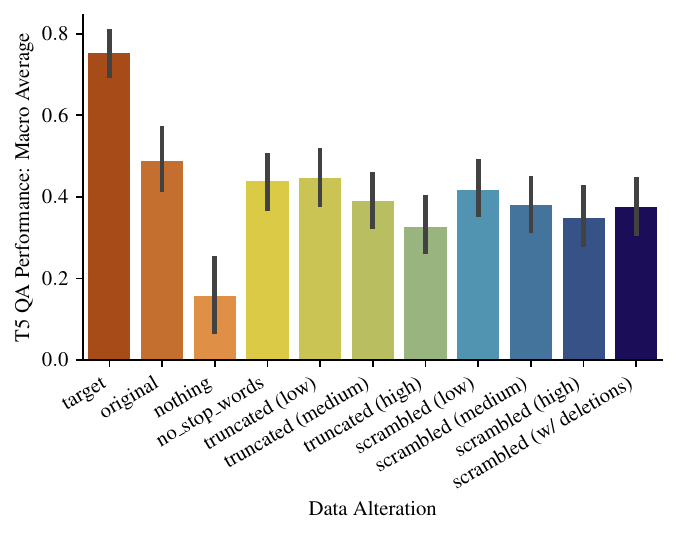}
        \vspace{-2em}
        \caption{Results from the feasibility study. $y$-axis shows macro average performance over the 17 datasets from UnifiedQA \citep{khashabi2022unifiedqav2}.}
        \label{fig:feasibility}
\end{wrapfigure}

\begin{figure}[t]
\begin{center}
\includegraphics[width=\columnwidth]{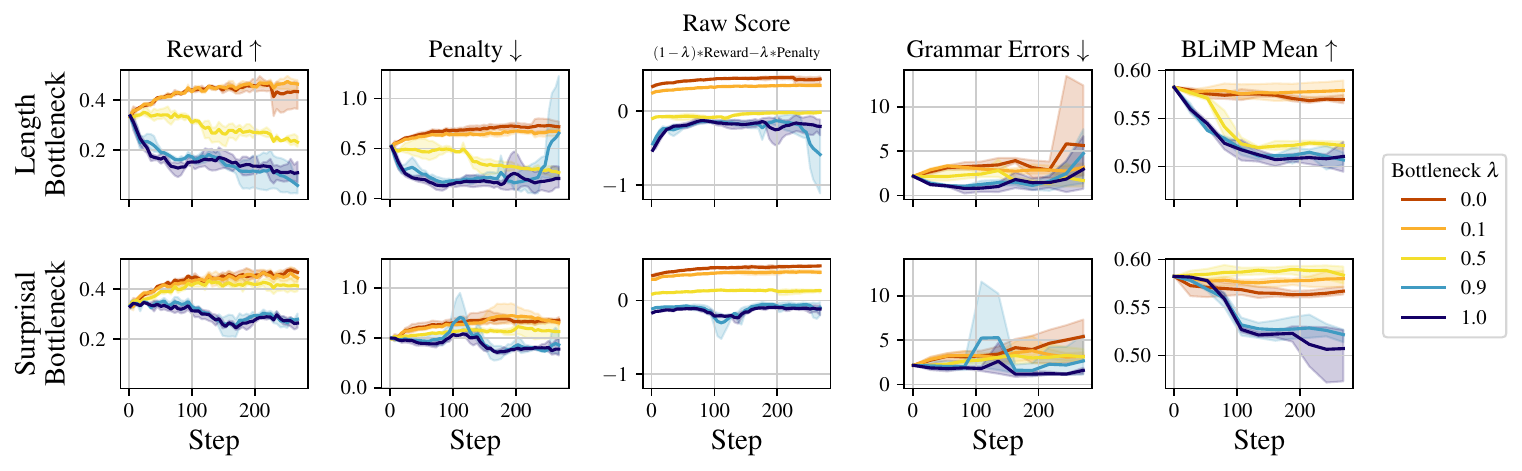}
\includegraphics[width=\textwidth]{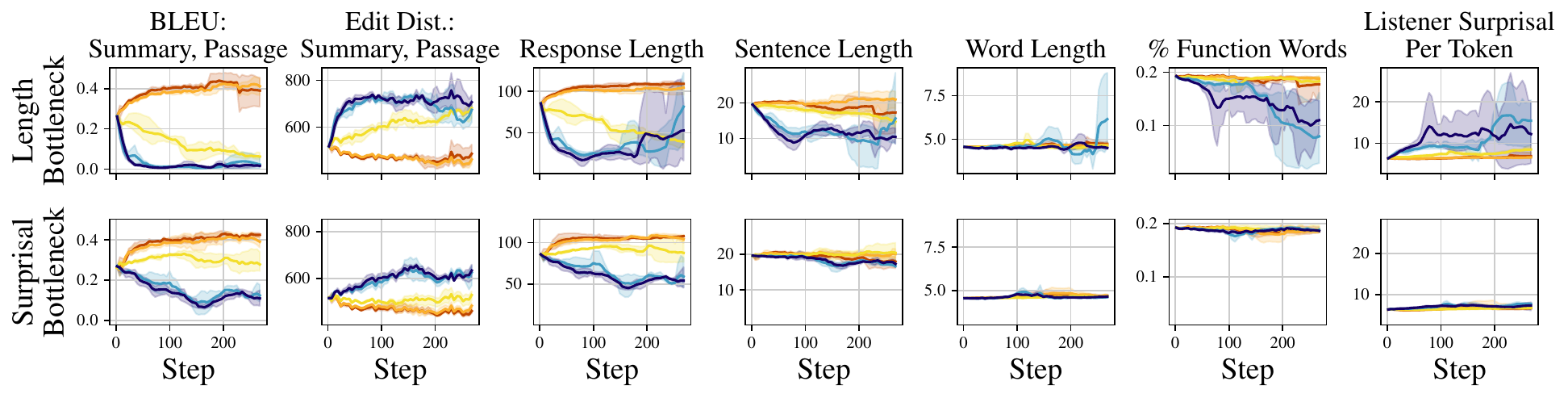}
\end{center}
\caption{The trajectory of reward, penalty, and score, as well as grammatical performance during training. We plot the mean value and 95\% confidence interval for all trajectories. BLiMP is a macro-average over all BLiMP categories. Grammar errors were estimated using LanguageTool. Below are textual properties of the summaries produced by the speaker throughout training.}
\label{rewardNpenalty}
\label{fig:grammaticality}
\label{fig:text-properties}
\end{figure}


Our objective is mainly to provide a proof of concept that an abstract reference game provides a learning signal for language acquisition in LMs.
To validate the viability of our setup in principle, we conduct a feasibility study measuring how the response of the listener model changes as a function of the quality of the passage it is provided.
To enable language acquisition, the reward signal should correlate with grammaticality, i.e., if the speaker produces ungrammatical summaries, the listener's performance and therefore the reward should degrade.
We expect this to be the case as we use a frozen pretrained QA model as the listener, and ungrammatical texts are out-of-domain.
Furthermore, to enable stable reinforcement learning, the reward signal should increase smoothly as the quality of the summaries improves.
We test perturbations of the context on the test and validation datasets used by \citet{khashabi2022unifiedqav2} to train UnifiedQA.%
\footnote{For abstractive datasets, listener performance is measured as ROUGE score between the generated answer and the ground-truth. 
For multiple-choice datasets performance is measured as accuracy.}
Results are provided in Figure \ref{fig:feasibility}.

\paragraph{Skylines/Baseline} As extreme baselines and skylines, we provide as context (1) the ground truth answer verbatim (target), (2) the original passage in its entirety, or (3) nothing.
Unsurprisingly, performance is best given the target, second best given the original passage, and worst given nothing. 
We note that performance given the target is not at ceiling, likely because these texts are out-of-domain relative to the typical passage that the UnifiedQA listener model was trained on.

\paragraph{Deletion}
We experiment with deletions and truncation of portions of the passage.
This aims to test how much the listener's ability to answer the questions depends on the completeness of the context. 
Since we hypothesized that a strict bottleneck could result in telegraphic speech, we first test how the model responds to a version of the context that has all stop words removed. 
We discover that there is a minor loss in performance compared to the original passage.
This confirms that there is some incentive to produce grammatical summaries, even in the presence of a length bottleneck.

Next, we truncate a fraction of context by picking a random starting point $s$ and then removing the next $n$ tokens. 
$n$ is drawn from the uniform distribution $\mathcal{U}_{[0,c(|\text{context}|-s)]}$, where $c$ dictates how much context is deleted. 
We repeat this process $m$ times.
We vary $m$ and $n$ to produce high, medium, and low degrees of truncation.
Figure \ref{fig:feasibility} shows that the listener performance is inversely correlated to the amount of data omitted. 
Furthermore, this relation seems to be gradual across all datasets. 

\paragraph{Permutation}
We simulate word order anomalies in the speaker output by scrambling parts of the context. 
We permute words within the sentence boundaries, preserving the overall order of sentences.
Specifically, every pair of words $(i,j)$ in a sentence has a probability $s$ of being switched.
We vary $s$ to produce low, medium, and high degrees of scrambling.
This process affects the syntax as well as the semantics of the sentence, e.g. by switching the subject and object. 
In addition, we perform one experiment where each word has a probability $d$ of being deleted. 
The results are consistent with the omission test, showing that performance degrades as a function of the degree of corruption.

\section{Experiment 2: Training from Scratch}\label{sec:training-from-scratch}
To most closely emulate human language acquisition we first experiment with a randomly initialized language model. 
However, we predicted that this setting would be challenging, as reinforcement learning is typically only applied to models with substantial pretraining \citep{ouyangTrainingLanguageModels2022}.
In order to increase the stability of optimization on the summarization game, we begin by training on next-word prediction for several epochs, before alternating between next-word prediction and summarization game objectives.
We also eschew the use of any bottleneck in these experiments in order to maximize the chance of learning a policy that increases the reward.
We conduct four runs, and track the reward trajectory (see Figure \ref{fig:from_scratch} in the Appendix). 
Despite our efforts to increase the success of reinforcement learning in this setting, the reward does not increase in any of our experiments.
Furthermore, the models quickly degenerate into nonsensical output.\footnote{In exploratory runs, we also unsuccessfully attempt to encourage the learning of a summarization policy by interspersing training as describe above with periods of log-likelihood-based summarization training on the CNN Daily Mail dataset \citep{see-etal-2017-get}.} 
As we have established the feasibility of the reward signal in Section \ref{sec:feasibility} but fail to observe successful optimization of the reward when training from scratch, our next experiment further advantages the learner model.

\section{Experiment 3: Fine-Tuning}\label{sec:fine-tuning}
Rather than train from scratch, we start with a pretrained foundation model and train only on the summarization game. 
We systematically sweep over 5 values for the bottleneck strength hyperparameter $\lambda\in\{0, 0.1, 0.5, 0.9, 1\}$ for both bottleneck variants. 
To account for the stochastic nature of training, we repeat each configuration with three different random seeds (see \cref{tab:hyperparameters}).

\paragraph{Task Performance}
At $\lambda=0$ the bottleneck is disabled. 
We observe that over time the speaker learns to produce better summaries and increase its reward (cf. Fig. \ref{rewardNpenalty}). 
However, it does so at the cost of becoming more verbose.
The penalty for both bottlenecks increase from about 0.5 at the start of training (i.e., summaries are about 50\% the length/surprisal of the original passage) to above 0.7 (summaries are 70\% the length/surprisal of the original passage).
This suggests that the models may be trending towards the trivial strategy described in \cref{sec:bottleneck}.
However, no model actually converges on that strategy even with $\lambda=0$.%

As $\lambda$ increases and more weight is assigned to the penalty, reward and penalty both begin to decrease.
Once a certain threshold of $\lambda$ is crossed, the penalty is favored to such an extent that reward and penalty start to decrease below their initial value.
For the length-based bottleneck that point is at $\lambda\geq0.5$, whereas for the surprisal bottleneck it happens at $\lambda\geq0.9$.
This suggests that the surprisal bottleneck is more consistent with high-quality summaries than the length bottleneck, as reward can remain relatively high even when more weight is put on reducing the penalty.


\paragraph{Language Drift}

To understand how the summaries change throughout training we track an array of text metrics (cf. Fig. \ref{fig:text-properties}). 
We notice that for runs that increase the reward the BLEU score between the original passage and the summary increases and the edit distance decreases---i.e., they become more alike---and the summary length increases. 
For runs with decreasing reward, this pattern is reversed.%
\footnote{As BLEU and Levenshtein distance are sensitive to the difference in length of the texts, it is not possible to tell how much of the change in similarity in these metrics is merely due to length variations.} 
Yet, the model never learns to copy the entire passage verbatim.%
\footnote{In preliminary experiments one model adopted this strategy.}
Examples of generated texts are given in Table \ref{tab:bottleneck-examples} in the Appendix.

In the case of the length bottleneck, the fraction of function words, sentence length, listener surprisal, and word length all paint a very consistent picture: 
The more weight is placed on the penalty the more content-dense and telegraphic the summaries become.
For the surprisal-based bottleneck, this effect is significantly reduced. 
Only sentence length and surprisal show minor changes.

Finally, we investigate the effect training had on grammaticality. 
We evaluate grammatical errors using the LanguageTool grammatical error detector,%
\footnote{\href{https://languagetool.org/}{https://languagetool.org/}} 
which counts the number of grammatical errors in each generated summary,
and BLiMP \citep{warstadt_blimp_2020}, which uses minimal pairs to test LMs' fine-grained grammatical knowledge in a zero-shot setting.
We do not observe any improvement in either measure.%
\footnote{BLiMP scores are close to chance (0.5), even before fine-tuning.
While this may be surprising considering T5's large training set and effectiveness as a foundation model, similarly scores were reported by \citet{babylm}.
A different pretrained model would likely perform substantially better on BLiMP in our setting.}

\section{Discussion and Conclusion}

Our findings point both to the potential of the abstract reference game that we introduce, as well as the difficulty of training current LMs in this setting.
While the feasibility study (\cref{sec:feasibility}) suggests that it is possible to extract a learning signal for syntax and semantics from this form of communicative feedback, our experiments fail to do so.
There are many explanations for why this might be the case, while still maintaining the possibility of getting strong results with minimal alterations.
Future work should explore broader hyperparameter searches, longer training times (we train for 240 steps), larger QA training sets (ours is $\sim$500k tokens), and model architectures besides T5.

Although the summarization game does not improve grammatical performance, we find that it does elicit meaningful changes in the language use of the speaker as a function of the learning signal.
Namely, we observe the use of copying with no bottleneck, and the use of telegraphic language with a length bottleneck.
Even without seeking to increase grammatical performance, there is interesting work to do exploring speaker and listener costs.
For instance, children are known to go through 1- and 2-word utterance phases \citep{bloom1970language} during development. 
This could be implemented in our framework as a curriculum over length bottlenecks. 
Under such a constraint, we might expect to see more human-like word-learning trajectories:
While language models tend to learn function words earlier \citep{chang2022word,ficarra2025distributional}, children prioritize content words, much like the speaker models under a length bottleneck.

There is a wide range of other variations that can be explored.
For instance, a speaker-surprisal bottleneck can model production difficulty.
Providing both the speaker and listener with part of the original passage can help to learn theory of mind and model pragmatic effects arising from common ground \citep{stalnakerAssertion1978}.
Allowing for multiple turns in a single exchange would increase the realism and provide another way to operationalize communicative feedback.
While our contribution is not the first to explore the area of interactive learning, it represents an initial step down a new path toward providing LMs with communicative feedback, and, perhaps just as importantly, communicative intent and a functional relationship with language.

\bibliographystyle{colm2025_conference}

\appendix
\section*{Training from Scratch}\label{sec:training-from-scratch-examples}
\begin{figure}[H]
\centering
\includegraphics[width=0.7\linewidth]{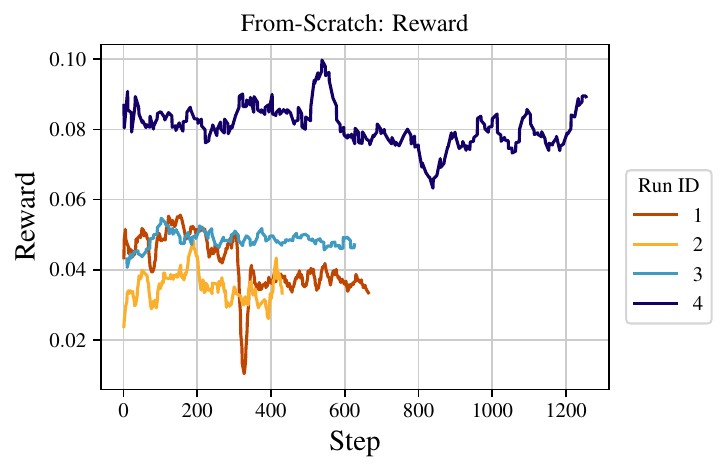}
\caption{Example reward trajectories in models trained from scratch, with rolling average window of 10 steps. Runs were terminated once the model had degenerated into nonsensical output. Trajectories start after initial pretraining on next-word prediction.}\label{fig:from_scratch}
\end{figure}

\section*{Hyperparameters}
Prior to running the experiments reported in Section \ref{sec:fine-tuning}, we conducted a Bayesian hyperparameter search using the Weights \& Biases Sweeps API \citep{biewald2020experiment} with the following search ranges: 

\begin{table}[h]
    \centering
    \begin{tabular}{lrr}
    \toprule
        \textbf{Hyperparameter} & \textbf{Min} & \textbf{Max}\\
    \midrule
        Learning Rate & 1e-6 & 1e-3\\
        Temperature & 0.5 & 1.5\\
        Top-$k$ & 10 & 1000\\
        Top-$p$ & 0.01 & 0.20\\
        \# Beams & 3 & 30\\
    \bottomrule
    \end{tabular}
    \caption{Search ranges for the initial hyperparameter sweep.}
    \label{tab:hp_sweep}
\end{table}

Based on the results of the sweep, we select the single best configuration, which we use in all the experiments reported in Section \ref{sec:fine-tuning}, varying only random seed. These and all other hyperparameters are as follows:

\begin{table}[h]
    \centering
    \label{tab:hyperparams}
    \begin{tabular}{lr}
    \toprule
    \textbf{Hyperparameter} & \textbf{Value} \\
    \midrule
    Learning Rate & 1.41e-5 \\
    Ratio Threshold & 10.0 \\
    PPO Epochs & 4 \\
    PPO Mini Batch Size & 64 \\
    PPO Batch Size & 512 \\
    Minimum Length & 2 \\
    Top-$k$ & 50 \\
    Top-$p$ & 0.1 \\
    Num. Beams & 3 \\
    Epsilon Cut-off & 3e-3 \\
    Max. New Tokens & 200 \\
    Use Score Norm. & True \\
    Use Score Scaling & True \\
    Steps & 240 \\
    \bottomrule
    \end{tabular}
    \caption{Training hyperparameters for the bottleneck experiments. Each configuration is run on three different seeds (7, 42, 99). Parameters not reported use TRL's default values.}\label{tab:hyperparameters}
\end{table}

\begin{table*}[t]
    \centering
    \begin{tabular}{cp{0.75\linewidth}c}
    \toprule
    \multicolumn{3}{p{\linewidth}}
    {\textbf{Query:} In 2008, Japan Aerospace Exploration Agency's SELENE probe observed evidence of the halo surrounding the Apollo 15 Lunar Module blast crater while orbiting above the lunar surface. In 2009, NASA's robotic Lunar Reconnaissance Orbiter, while orbiting 50 kilometers (31 mi) above the Moon, began photographing the remnants of the Apollo program left on the lunar surface, and photographed each site where manned Apollo flights landed. All of the U. S. flags left on the Moon during the Apollo missions were found to still be standing, with the exception of the one left during the Apollo 11 mission, which was blown over during that mission's lift-off from the lunar surface and return to the mission Command Module in lunar orbit; the degree to which these flags retain their original colors remains unknown.} \\
    \addlinespace[1em]
    \midrule
    \multicolumn{3}{c}{\textbf{Surprisal Bottleneck}} \\
    \midrule
    \textbf{$\lambda$} & \textbf{Response} & \textbf{Reward} \\
    \midrule
    0.0 & Flags left on the Moon during the Apollo 11 missions were found to be standing. In 2008, Japan Aerospace Exploration Agency probe observed evidence of the halo surrounding the blast halo while orbiting above the lunar surface. In 2009, NASA's Lunar Reconnaissance Orbiter began photographing the remnants of the Apollo program left on the lunar surface. All of the flags left on the Moon during the Apollo missions were found to still be standing, with the exception of the one left during the Apollo. & 0.6154 \\ \addlinespace[1em]
    0.5 & The robotic Lunar Reconnaissance Orbiter landed 50 kilometers (31 mi) above the Moon in 2009. In 2009, NASA's robotic Lunar Reconnaissance Orbiter began photographing each site where manned Apollo flights landed. All of the U. S. flags left on Moon during the Apollo 11 mission were found to still standing. & 0.3716 \\ \addlinespace[1em]
    1.0 & SELENE's SELENE probe in 2009. In 2009, NASA's robotic Lunar Reconnaissance Orbiter began photographing the remnants of the Apollo missions. & 0.3285 \\[3ex]
    \addlinespace[0.5em]
    \midrule
    \multicolumn{3}{c}{\textbf{Length Bottleneck}} \\
    \midrule
    \textbf{$\lambda$} & \textbf{Response} & \textbf{Reward} \\
    \midrule
    0.0 & In 2008, NASA's robotic Lunar Reconnaissance Orbiter began photographing the remnants of the Apollo 11 missions, and photographed each site where manned Apollo flights landed. In 2008, NASA's robotic Lunar Reconnaissance Orbiter began photographing the remnants left on the Moon, and photographed each site where manned Apollo flights landed. In 2008, NASA's robotic Lunar Reconnaissance Orbiter, while orbiting 50 km (31 mi) above the Moon, began photograph. & 0.6000 \\ \addlinespace[1em]
    0.5 &  (31 mi) above the Moon. & 0.1085 \\ \addlinespace[1em]
    1.0 & Exploration Agency's SELENE probe observed evidence of the halo surrounding the Apollo 15 Lunar Module blast crater. & -0.1710
    \\[3ex]
    \bottomrule
    \end{tabular}
    \caption{Example query-response pairs with different bottleneck types and strengths ($\lambda$). The responses show how different bottleneck penalties affect the model output.}
    \label{tab:bottleneck-examples}
\end{table*}

\end{document}